%% file: acl_latex.tex
\pdfoutput=1

\documentclass[11pt]{article}

\usepackage[preprint]{acl}

\usepackage{times}
\usepackage{latexsym}

\usepackage[T1]{fontenc}

\usepackage[utf8]{inputenc}

\usepackage{microtype}

\usepackage{inconsolata}

\usepackage{bm}
\usepackage{color}
\usepackage{graphicx}
\usepackage{amsmath}
\usepackage{amsthm}
\usepackage{xcolor}
\usepackage{multirow}
\usepackage{booktabs}
\usepackage{rotating}

\title{On LLMs-Driven Synthetic Data Generation, Curation, \\
and Evaluation: A Survey}

\author{Lin Long$^1$, Rui Wang$^1$, Ruixuan Xiao$^1$ \\
{\bf Junbo Zhao$^1$, Xiao Ding$^2$, Gang Chen$^1$, Haobo Wang$^1$\thanks{Corresponding author.}} \\
$^1$Zhejiang University, China \quad $^2$Harbin 
  Institute of Technology, China \\
  {\small \textit{Correspondence}: \texttt{wanghaobo@zju.edu.cn}}}

\begin{document}
\maketitle

\input{sec/0_abstract}
\input{sec/1_introduction}

\input{sec/2_preliminaries}
\input{sec/3_data_generation}
\input{sec/4_data_curation}

\input{sec/5_data_evaluation}
\input{sec/7_discussion_conclusion}

\input{sec/limitations}

\bibliography{anthology,custom}

\input{sec/appendix}

\end{document}

%% file: sec/0_abstract.tex
\begin{abstract}
Within the evolving landscape of deep learning, the dilemma of data quantity and quality has been a long-standing problem. The recent advent of Large Language Models (LLMs) offers a data-centric solution to alleviate the limitations of real-world data with synthetic data generation. However, current investigations into this field lack a unified framework and mostly stay on the surface. Therefore, this paper provides an organization of relevant studies based on a generic workflow of synthetic data generation. By doing so, we highlight the gaps within existing research and outline prospective avenues for future study. This work aims to shepherd the academic and industrial communities towards deeper, more methodical inquiries into the capabilities and applications of LLMs-driven synthetic data generation.
\end{abstract}

%% file: sec/1_introduction.tex
\section{Introduction}
The game-changing emergence of Large Language Models (LLMs) instigated a significant paradigm shift in the field of deep learning~\cite{abs-2303-11717,abs-2301-07597,abs-2302-04023}. 
Despite these advancements, a large amount of high-quality data remains the foundation for building robust NLP models~\cite{abs-2404-14361}. To be more specific, here high-quality data typically refers to diverse data that carries rich supervision signals (generally in the form of labels) closely aligned with human intent.
However, fulfilling such data reliance with human data can be challenging or even unrealistic sometimes, due to high costs, data scarcity, privacy concerns, etc. \cite{kurakin2023harnessing}. Moreover, several studies~\cite{abs-2309-16349,abs-2312-06585,abs-2303-15056} have highlighted that human-generated data, being inherently susceptible to biases and errors, may not even be optimal for model training or evaluation. These considerations necessitate a more serious inquiry into the question: are there other more \textit{effective and scalable} methods of data collection that can overcome the current limitations?


Given the recent advancements in LLMs, which demonstrate the capability to generate fluent text on par with human output~\cite{hartvigsen-etal-2022-toxigen, sahu-etal-2022-data, ye-etal-2022-zerogen, abs-2303-04360, GaoPLXY0ZLLK23}, synthetic data produced by LLMs emerges as a viable alternative or supplement to human-generated data. Specifically, synthetic data is designed to mimic the characteristics and patterns of real-world data~\cite{abs-2404-07503}. On the one hand, LLMs, through extensive pretraining, have acquired a vast repository of knowledge and demonstrate exceptional linguistic comprehension~\cite{abs-2207-02516,ding-etal-2023-gpt}, which forms a foundation for generating faithful data. On the other hand, the profound instruction-following capabilities of LLMs allow better controllability and adaptability over the generation process, facilitating the creation of tailored datasets for specific applications with more flexible process designs~\cite{abs-2305-07759}. These two advantages make LLMs highly promising synthetic data generators.

\begin{figure}
\centering
\includegraphics[width=1\linewidth]{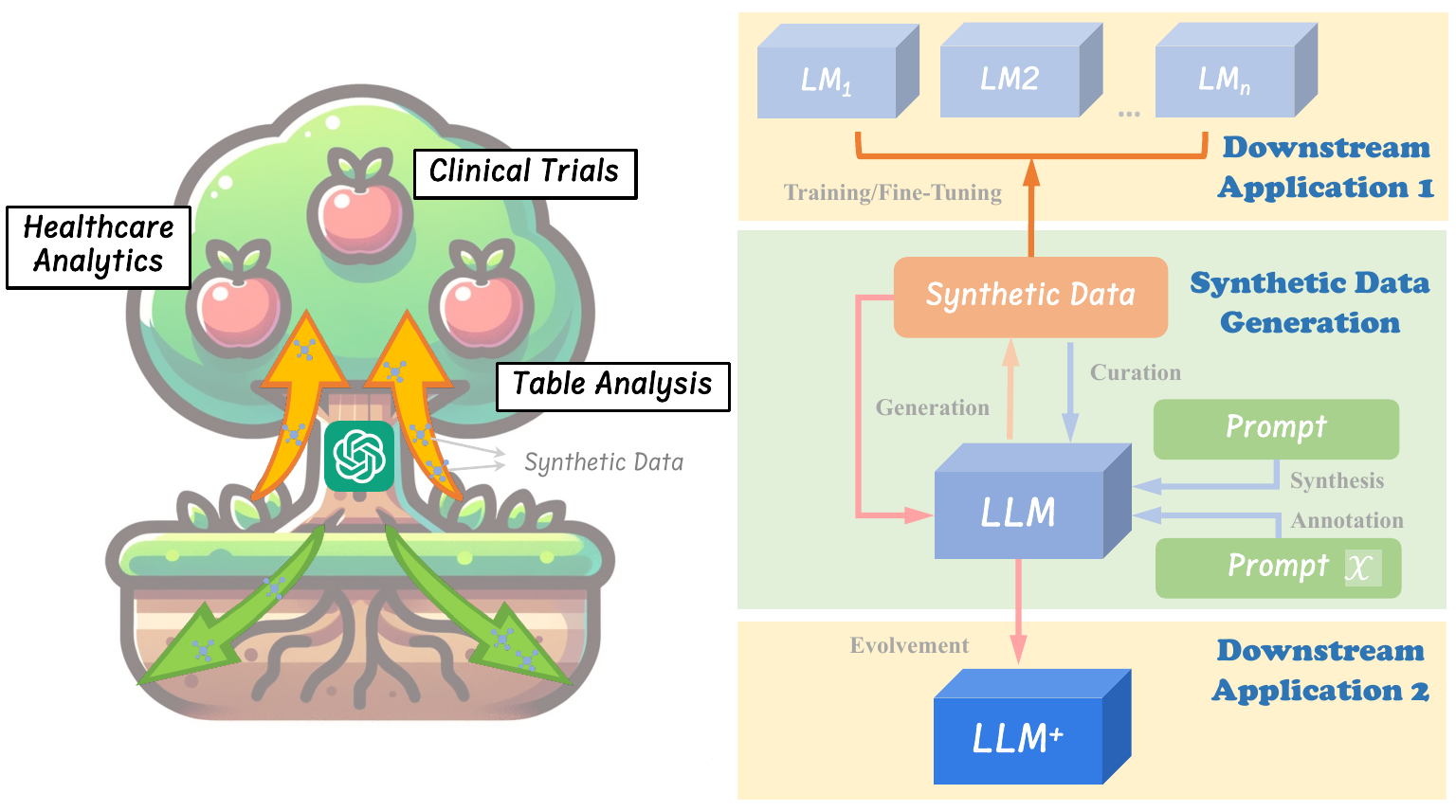}
\caption{Illustration of the LLMs-based application ecosystem, where synthetic data serves as the flowing nutrients for fruiting (training of small LMs or fine-tuning task-specific LLMs) and rooting (training stronger LLMs or self-improvement).}
\label{fig:illustration}
\end{figure}


As a pivotal application of LLMs, synthetic data generation holds significant importance for the development of deep learning. As shown in Figure~\ref{fig:illustration}, 
LLMs-driven synthetic data generation~\cite{LiZL023,wang-etal-2021-want-reduce,seedat2023curated} enables the automation of the entire model training and evaluation process with minimal human participation required in the loop~\cite{0001GHW00023}, which allows the advantages of deep learning models to be applied across a broader range of applications. Beyond providing a scalable supply of training and testing data, LLM-driven synthetic data generation also may pave the way for developing next-generation LLMs. Insights from TinyStories~\cite{abs-2305-07759} and the \textit{Phi} series~\cite{abs-2306-11644,abs-2309-05463} emphasize that data quality is crucial for effective model learning, while LLMs empower us to actively ``design'' what the models learn through data manipulation, significantly enhancing the efficacy and controllability of model training. 
As of June 2024, there are over $300$ datasets on Hugging Face\footnote{\url https://huggingface.co} that are tagged as ``synthetic'', with many mainstream LLMs leveraging high-quality synthetic data for training, including Alpaca~\cite{alpaca}, Vicuna~\cite{zheng2023judging}, OpenHermes 2.5, and Openchat 3.5~\cite{wang2023openchat}.

Though seemingly straightforward, 
generating synthetic datasets that simultaneously have high correctness and sufficient diversity requires careful process designs and involves a lot of tricks~\cite{abs-2404-14361},
making LLMs-driven synthetic data generation a non-trivial problem. While most existing works generally target data generation for various tasks (e.g., pre-training~\cite{abs-2306-11644,abs-2309-05463,abs-2305-07759}, fine-tuning~\cite{abs-2306-02707,abs-2311-11045,abs-2304-12244}, evaluation~\cite{feng-etal-2023-factkb,abs-2403-18802}) across different domains (e.g., math~\cite{abs-2309-12284,abs-2308-09583}, code~\cite{abs-2306-08568,abs-2312-02120}, instruction~\cite{honovich-etal-2023-unnatural,wang-etal-2023-self-instruct}), they share many common ideas. To address the lack of a unified framework in the emerging field of LLM-driven synthetic data generation and develop a general workflow, this survey investigates recent studies and organizes them according to the topics of generation, curation, and evaluation, which are closely related, as shown in Figure~\ref{fig:taxonomy}. Our primary aim is to provide a comprehensive overview of the current state of the field, identify key areas of focus, and highlight the gaps that remain to be addressed. We hope to bring insights to both the academic and industrial communities and drive further development in LLM-driven synthetic data generation.

%% file: sec/2_preliminaries.tex
\section{Preliminaries}
\subsection{Problem Definition}
In this paper, we investigate the challenge of generating high-quality synthetic data using pre-trained LLMs, denoted as $\mathcal{M}$. Rather than creating new datasets from scratch, in more cases, we perform data augmentation with a small number of seed samples or unlabeled inputs, which we denote uniformly as $\mathcal{D}_\text{sup}$. Although optional for LLMs-driven synthetic data generation, $\mathcal{D}_\text{sup}$ can typically provide valuable supporting information when available. Consequently, the overall generation task can be formulated as:
\begin{equation}
    \mathcal{D}_\text{gen} \leftarrow \mathcal{M}_{p}(\mathcal{T}, \mathcal{D}_\text{sup})\text{,}
    \label{eq:overall}
\end{equation}
where $\mathcal{D}_\text{gen}$ represents the final generated dataset, and $p$ refers to the prompt used for model inference. $\mathcal{T}$ specifies the generation task, such as rewriting, question answering,  annotation, etc. Notably, data annotation as a specialized paradigm of synthetic data generation, has particularly extensive applicability, including RLAIF~\cite{abs-2212-08073} and LLMs-based evaluation~\cite{abs-2307-08701,zheng2023judging,abs-2310-08491}, which may involve specific challenges and corresponding solution techniques. Due to page limitations, further details about data annotation can be found in Appendix~\ref{app:data_annotation}.

\begin{figure}[!t]
\centering
\includegraphics[width=1\linewidth]{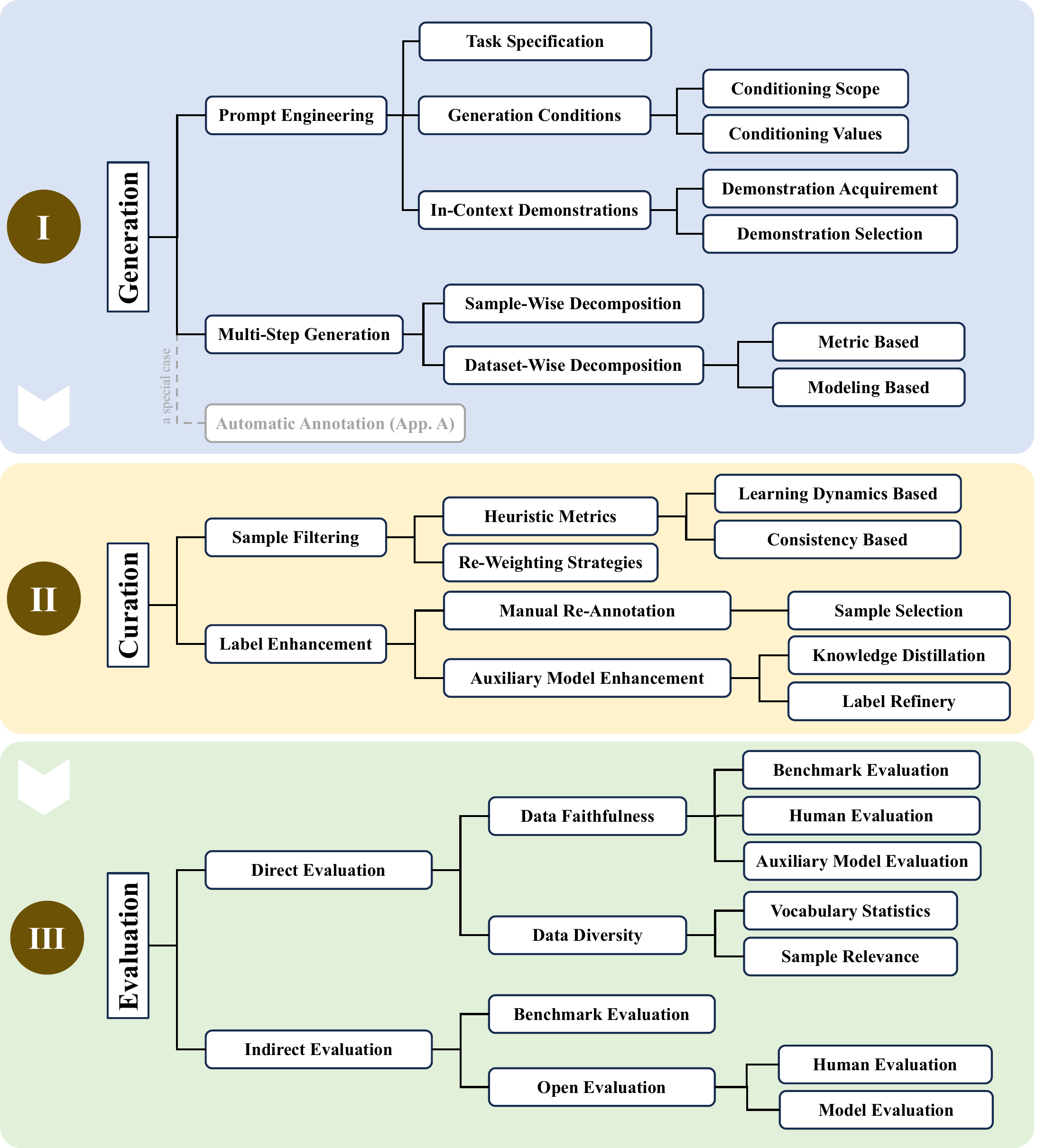}
\caption{A taxonomy of LLMs-driven synthetic data generation, curation, and evaluation.}
\label{fig:taxonomy}
\end{figure}

\subsection{Requirements of $\mathcal{D}_\text{gen}$} 
\label{sec:requirements}
Briefly speaking, our goal is to generate data that closely aligns with evaluation metrics. While the standard of high-quality data may vary across different downstream tasks, there are two general requirements that are considered challenging in most existing literature:
\begin{itemize}
    \item \textbf{Faithfulness.}
    To provide valid supervision, the generated data must first be logically and grammatically coherent. However, the inherent problems of hallucination fat-tailed knowledge distribution of LLMs can introduce significant noise into the generated results, manifesting as factual errors, incorrect labels, or irrelevant content. These issues become more pronounced when generating long, complex, or domain-specific data.
    \item \textbf{Diversity.}
    Diversity captures the variation among the generated data, reflecting differences in text length, topic, or even writing style. It is crucial for generating synthetic samples that mimic the diversified nature of real-world data, thereby preventing overfitting and bias during model training or evaluation. Nevertheless, due to the inherent biases of LLMs, uncontrolled generated content often tends to be monotonous, limiting its applicability in downstream tasks.
\end{itemize}
These two requirements are the focal points of most current research efforts. In the subsequent workflow, we will introduce how different methods address these issues.

%% file: sec/3_data_generation.tex
\definecolor{yellow_highlight}{RGB}{249,218,120}
\definecolor{blue_highlight}{RGB}{224,235,246}
\definecolor{red_highlight}{RGB}{245,191,196}

\begin{figure*}[!ht]
\centering
\includegraphics[width=1\textwidth]{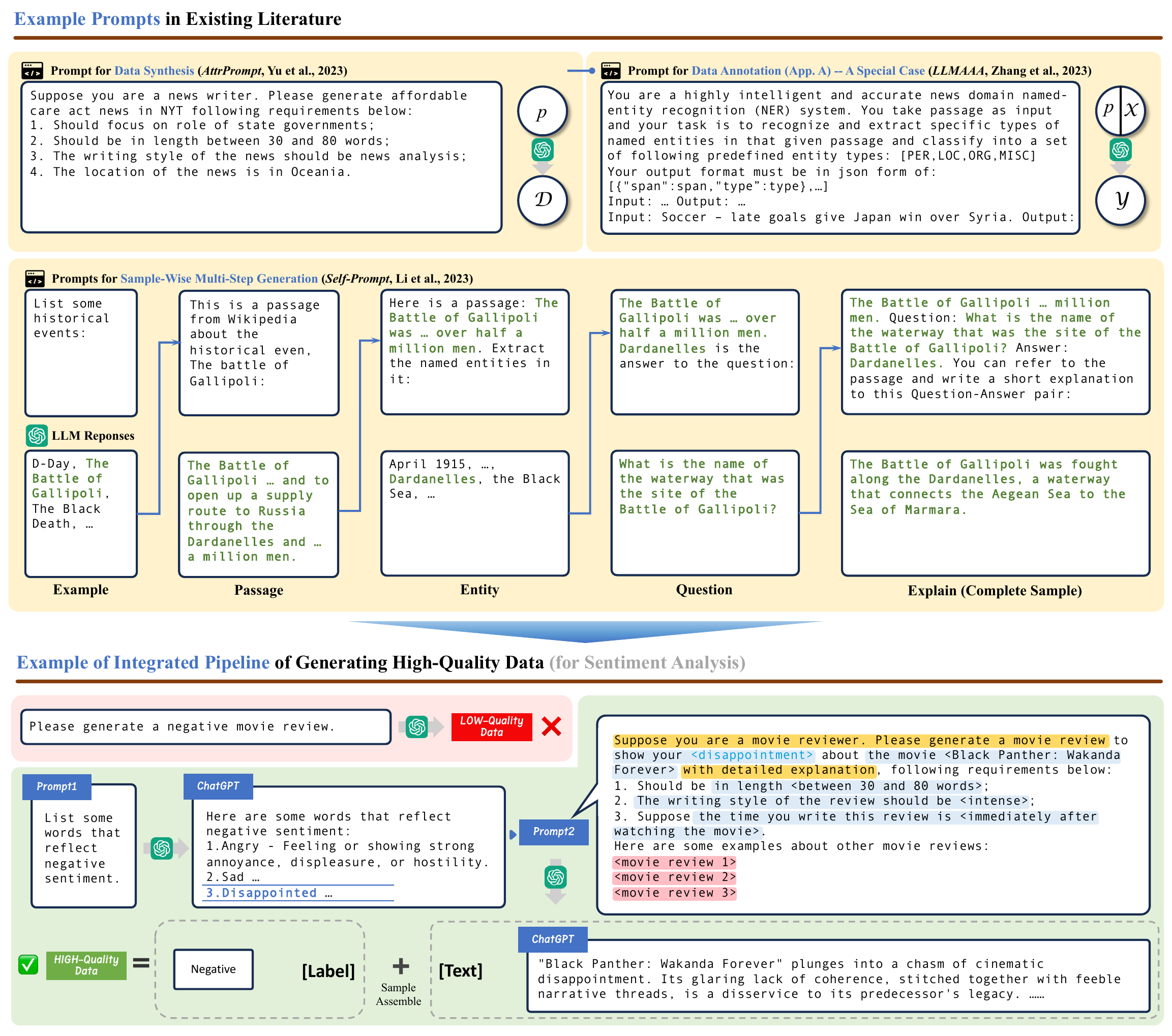}
\caption{A toy example of effective synthetic data generation. The corresponding fields for \colorbox{yellow_highlight}{task specification}, \colorbox{blue_highlight}{conditions}, and \colorbox{red_highlight}{in-context demonstrations} are highlighted, while < > marks the switchable contents.}
\label{fig:prompt}
\end{figure*}

\section{Generic Workflow}

Existing studies on LLMs-driven synthetic data generation generally incorporate three main topics: generation, curation, and evaluation. Various approaches are employed within these aspects to collaboratively achieve optimal data generation.

\subsection{Data Generation}
In this section, we systematically summarize some common practices for synthetic data generation with LLMs, which can be roughly divided into prompt engineering and multi-step generation. An overall illustration is provided in Figure~\ref{fig:prompt}.

\subsubsection{Prompt Engineering}
\label{subsubsec:prompt_engineering}
\label{subsec:prompt_engineering}
One of the greatest advantages of LLMs for synthetic data generation is their instruction-following capability, which contributes to great controllability \cite{Wang2023PandaLMAA,Radford2019LanguageMA}. Therefore, many approaches try to guide LLMs with heuristic prompts to enhance the faithfulness and diversity of the synthetic data~\cite{abs-2404-07503}. 

Empirically, an effective prompt generally contains three key elements: \textit{task specification} $e_\text{task}$, \textit{generation conditions} $e_\text{condition}$, and \textit{in-context demonstrations} $e_\text{demo}$, which are then collectively wrapped with a template $E$ into the form of natural instruction:
\begin{equation}
    p(\mathcal{T},\mathcal{D}) \leftarrow E(e_\text{task},e_\text{condition},e_\text{demo})\text{.}
    \label{eq:instruction}
\end{equation}
As shown above, both the generation task $\mathcal{T}$ and the support dataset $\mathcal{D}$ will affect the design of $p$. Next, we will proceed to detail how each part of the prompt should be appropriately designed to accommodate various scenarios.

\paragraph{Task Specification.}
In traditional crowdsourced annotation scenarios, the recruited workers are commonly offered a codebook that specifies the necessary contexts, such as task purpose, data explanation, and other background knowledge, so that they can better understand their jobs \cite{abs-2303-15056}. Similarly, such task specification is crucial for setting the right context for LLMs-driven data generation, which can also include role-play~\cite{LiZL023}, format clarification, knowledge augmentation~\cite{abs-2311-00287,abs-2403-01081}, etc. Evidence shows that a simple prologue such as ``suppose you are a \texttt{\{xxx\}}'' can significantly improve the LLMs' performance by setting up a proper scenario for data generation and allowing the LLMs to better take on the roles \cite{LiZL023}. More formally, \citet{YooPKLP21} defines the task specification with a triplet of text type, label type, and label-token verbalizer. Such a description header is particularly important when extra domain expertise is demanded to address issues like terminology complexities in both context understanding and data generation. Consequently, \citet{abs-2311-00287} leverages external knowledge graphs and LLMs to obtain domain topics for context-informed prompting, which effectively enhances the faithfulness and complexity of generated data. 


\paragraph{Conditional Prompting.}
\label{para:conditional_prompting}
As mentioned in Section~\ref{sec:requirements}, a pivotal challenge in using LLMs for synthetic data generation is ensuring sufficient diversity, as directly prompting the LLMs to produce data for certain tasks often results in highly repetitive outputs, even with a high decoding temperature~\cite{abs-2404-14361,abs-2404-07503}. Addressing this problem, a widely adopted strategy is conditional prompting, which explicitly and concretely communicates to the LLMs the specific type of data desired.
The core of conditional prompting involves delineating the targeted data through the formulation of a series of condition-value pairs:
\begin{equation}
    e_\text{condition}=\{(c_1,v_1),(c_2,v_2),\cdots,(c_n,v_n)\}\text{,}
\end{equation}
which effectively characterizes the desired attributes and characteristics of the synthetic data.
With different combinations of such attributes, we can automatically achieve a degree of ``artificially defined'' diversity in the generated samples~\cite{abs-2306-11644,abs-2309-05463,abs-2305-07759}.
Conditional prompting not only allows better control over the diversity and coverage of the generated dataset but also refines the content to a narrower, more focused scope that is more likely to align with our specific expectations and requirements~\cite {LiZL023}. Current research on conditional prompting primarily centers on the following two subjects:
\begin{itemize}
    \item[1)] \textbf{Conditioning Scope.} As the backbone of $e_\text{condition}$, conditioning scope defined by $\{c_1,\cdots,c_n\}$ delineates the dimensions that we utilize to characterize our target data. Early studies \cite{GaoPLXY0ZLLK23,ye-etal-2022-zerogen,ye-etal-2022-progen} employed a basic output-conditional prompting strategy, utilizing the specific label associated with the classification task as the conditioning variable. The rationale behind this was primarily to maintain class balance and coverage. However, such a strategy is unsuitable for data lacking explicit category labels. Subsequent work by~\citet{abs-2306-15895} argues that conditional-prompting with finer-grained attributes (e.g., topics, length, and style \cite{abs-2311-00287}), can lead to more diversified generation due to the vast number of possible attribute combinations, being also applicable to open-ended data. Additionally, \citet{abs-2305-07759} also condition each generation on the task of incorporating three randomly chosen words into the generated story. This approach was also proven to significantly enhance the diversity of the generated data, shifting the focus from the heuristic features of the output to a more structured and targeted conditioning mechanism by adding ``creative randomness'' to the prompt~\cite{abs-2305-07759}.
    \item[2)] \textbf{Conditioning Values.} After defining the conditioning scope, we then need to assign concrete values to each condition. 
    Despite the seemingly straightforward strategy of sampling from the known classes or labels \cite{ye-etal-2022-zerogen}, there are cases where such an instance pool is unavailable. Addressing this problem, \citet{JosifoskiSP023} actively retrieves the conditioning instances from external knowledge graphs, while \citet{abs-2311-00287,ding-etal-2023-enhancing} leverage the LLMs to generate diversified instances for conditional prompting. Specifically, \citet{ding-etal-2023-enhancing} construct a concept tree to delve into different subtopics, ensuring the coverage of sampled conditioning values, which then contributes to more diverse generated data. Moreover, the prompt template $E$ can also be considered a special type of condition. It has been demonstrated that incorporating templates with a certain level of randomness throughout the generation process can enhance the diversity of the generated contents \cite{DBLP:conf/nips/MengHZH22}. 
\end{itemize}

\paragraph{In-Context Learning.}
\label{para:icl}
Due to the inherent bias of LLMs, it remains challenging to elicit favorable responses from the LLMs with merely task specification and conditional prompting. In this case, a straightforward yet effective strategy is to provide several demonstrations, which can serve as a form of implicit human guidance. Research has shown that, owing to LLMs' remarkable in-context learning (ICL) capabilities, a few exemplars can provide them with insights into the patterns exhibited in real-world data, thereby significantly improving the faithfulness of generated data \cite{LiZL023}. In the few-shot setting, where labeled samples are available in the support set $\mathcal{D}_\text{sup}$, these samples can be directly utilized as demonstrations for ICL. However, in scenarios where no ground truth data is available, approaches like Self-Instruct \cite{WangKMLSKH23} and Self-Prompting \cite{abs-2212-08635} instead leverage ICL with synthetic demonstrations generated by LLMs. This allows the models to learn from their own predictions or other teacher models, even in the absence of labeled data.

However, given the constraint of prompt length and data inconsistency, the quality of in-context samples significantly affects the effectiveness of in-context learning. \citet{abs-2403-01081} argue that randomly selecting in-context examples from the pool of seed samples, as done in Self-Instruct~\cite{WangKMLSKH23}, results in a lack of diversity and quality in the generated data. To address this issue, \citet{abs-2403-01081} opt for selecting examples that concentrate on specific aspects to better stimulate the long tail of knowledge inherent in LLMs. \citet{LiuSZDCC22} and \citet{SuKWSWX0OZS023} prioritize consistent samples as demonstrative examples based on their cosine similarity in the embedding space. Alternatively, \citet{ye-etal-2022-progen} selects the most informative samples using quantified influence scores to steer the generation process. To enhance the informativeness of in-context examples, \citet{abs-2303-16854} prompts LLMs to provide an explanation for each sample before integrating it into the prompt. This approach not only offers valuable additional information but also aligns well with the subsequent Chain-of-Thought generation.


\subsubsection{Multi-Step Generation}
In the previous paragraphs, we have introduced some common prompting strategies, which are typically designed for a specific generation task $\mathcal{T}$. However, in most cases, due to the lack of enough reasoning abilities, it is unrealistic to expect the LLMs to generate the entire desired dataset within a single reference, especially when targeting data with complex structures or semantics~\cite{abs-2310-04484}. 
In addressing this problem, a common strategy is multi-step generation, through which the overall generation process is manually decomposed into a chain of simpler sub-tasks $\mathcal{T}_{1:k}$, to force the LLMs to produce data in a step-by-step manner as scheduled:
\begin{equation}
\begin{aligned}
    \mathcal{D}_{i} \leftarrow \mathcal{M}^{i}_{p_i}(\mathcal{T}_{i},&\mathcal{D}_{0:i-1}), \ i=1,2,\cdots,k\text{,}
\end{aligned}
\label{eq:multi-step}
\end{equation}
where $\mathcal{D}_0=\mathcal{D}_\text{sup}$. Each intermediate output $\mathcal{D}_i$ is generated using model $\mathcal{M}^i$, prompted by $p_{i}$, for a sub-task $\mathcal{T}_{i}$. These outputs can then potentially be used in subsequent generations. By manually scheduling the generation procedure, we implicitly align the reasoning paths of LLMs with human prior knowledge.
Specifically, there are two common strategies for task decomposition: \textit{sample-wise} and \textit{dataset-wise} decomposition, which mainly aim at enhancing the quality of synthetic data at different scales.

\paragraph{Sample-Wise Decomposition.} A typical use-case of multi-step generation is for addressing the challenges of long-text processing and logical reasoning when dealing with multi-text data such as dialogues and entity-relation triplets. In such cases, a straightforward approach is to divide the sample into smaller chunks and generate only a portion of each sample at a time \cite{abs-2212-08635,ye-etal-2023-generating,WangKMLSKH23}. In this way, $\mathcal{D}_{1:k}$ can be considered as different parts of $\mathcal{D}_\text{gen}$:
\begin{equation}
    \mathcal{D}_\text{gen} = (\mathcal{D}_1,\mathcal{D}_2,\cdots,\mathcal{D}_k)\text{.}
\end{equation}
Notably, as shown in Eq.~\ref{eq:multi-step}, each iteration of the generation process can be conditioned on the previously generated contents. For example, \citet{ding-etal-2023-enhancing} prompts the LLMs to alternate between acting as the assistant and the user, replying to each other based on the context, ultimately producing a complete conversation transcript. In this way, the coherence among each internal component $\mathcal{D}_i$ can be pointedly reinforced with separated instructions, thus making it easier for the model to follow the requirements and generate more faithful data. It should be noted that $D_{1:k}$ may not necessarily form part of the final $D_\text{gen}$, instead, explicitly outputting some intermediate reasoning steps can also improve the generation of complex data \cite{abs-2212-08073,abs-2303-16854}. Chain-of-Thought (CoT) prompting stands out as one of the most popular strategies for improving the faithfulness of LLM-generated content~\cite{Wei0SBIXCLZ22}. Nevertheless, current research on the exploration of such latent metadata is still insufficient, leaving sample-wise task decomposition from a reasoning perspective an open problem for future studies.

\paragraph{Dataset-Wise Decomposition.}
In Section~\ref{para:conditional_prompting} we have introduced how to generate data with specified properties. However, generating a series of such data that can eventually form a dataset with good diversity and domain coverage requires long-term scheduling. To this end, dataset-wise task decomposition dynamically adjusts the conditions used at each stage of multi-step generation to ensure the overall dataset grows in the right direction:
\begin{equation}
    \mathcal{D}_\text{gen} = \bigcup_{i=1}^{k}\mathcal{D}_i\text{.}
\end{equation}
Specifically, S3 \cite{WangZS23} targets the most frequently mislabeled categories at each iteration, according to the performance of the downstream model trained on previously generated data. 
Similarly, \citet{HonovichSLS23,ShaoGSHDC23} utilize a generate-then-expand paradigm, to enhance the diversity of the overall dataset accordingly. Some other methods also leverage specific data structures to model the pathways of data generation. For example, Explore-Instruct \cite{WanHYQB023} models the domain space as a tree structure and continually refines the generated data along with tree traversal to promote both the specialization and domain coverage of the generated data. 

%% file: sec/4_data_curation.tex
\subsection{Data Curation}
\label{sec:data_curation}
After the preceding steps, one may excessively generate overflowing and theoretically unlimited data $\mathcal{D}_\text{gen}$. However, these datasets often comprise a considerable portion of noisy, worthless, or even toxic samples, which primarily stems from two causes. Firstly, LLMs can inevitably produce corrupted samples with incorrect labels due to the hallucination problem. Secondly, ineffective prompts containing ambiguous descriptions can trick the model into generating irrelevant or redundant samples. Consequently, directly utilizing these low-quality data without proper processing may have a significant negative impact. 

To address this, plenty of data curation approaches have been studied, which mainly fall into two dominant groups of \textit{high-quality sample filtering} and \textit{label enhancement} as elaborated below. 



\begin{figure}[!t]
\centering
\includegraphics[width=1\linewidth]{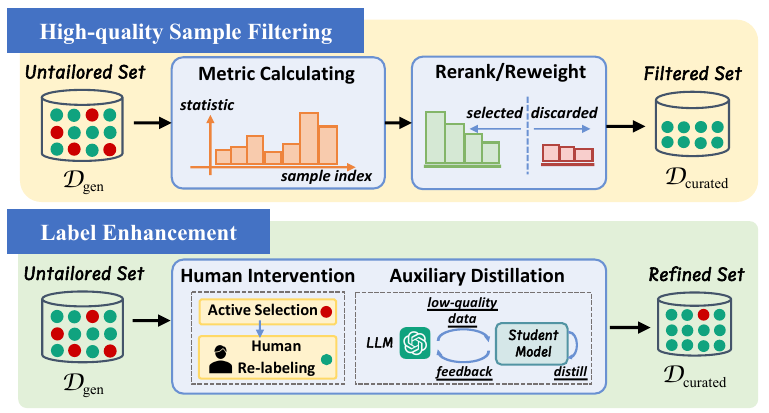}
\caption{Two dominant approaches of data curation.}
\label{fig:curation}
\end{figure}

\subsubsection{High-Quality Sample Filtering} 
Sample filtering aims to weed out undesired low-quality samples and obtain a more helpful subset $\mathcal{D}_\text{curated}\! \subset\!\mathcal{D}_\text{gen}$. These methods typically design \textit{heuristic criteria} or \textit{re-weighting functions} to rerank samples for filtering, as shown in Figure~\ref{fig:curation}.

\paragraph{Heuristic Metrics.}
For methods based on heuristic metrics, the key step is to design appropriate criteria based on the learning dynamics, such as confidence score \citep{seedat2023curated}, influence function \cite{ye-etal-2022-progen}, and generation ability \cite{DBLP:conf/nips/MengHZH22}. SuperGen \cite{DBLP:conf/nips/MengHZH22} employs the estimated generation probability to identify samples most related to the desired label. \citet{seedat2023curated} discard samples with both low confidence and low uncertainty. Some other methods assume that clean samples are prone to hold similar predictions under different conditions and employ cross-condition consistency for filtering. Specifically, such consistency can be between LLM and downstream classifier \cite{yu-etal-2023-regen}, between multiple executions \cite{ye-etal-2023-generating}, or between neighboring data points \cite{seedat2023curated}. \citet{abs-2307-08701} leverage the powerful text understanding capabilities of LLMs to assess the quality of different samples and filter out those with low scores. Results show that Alpagasus~\cite{abs-2307-08701}, trained on a much smaller but curated dataset, surpasses the original Alpaca~\cite{alpaca} across several benchmarks, underscoring the importance of data curation.


\paragraph{Sample Re-Weighting.}  
On the other hand, re-weighting methods believe all data are valuable but with varying importance.
Thus, they assign larger weights to correctly annotated or influential samples during downstream utilization \cite{zhang-etal-2023-llmaaa,GaoPLXY0ZLLK23,meng2023tuning}. For instance, SunGen \cite{GaoPLXY0ZLLK23} proposes an adaptive bi-level re-weighting algorithm without human annotations. FewGen \cite{meng2023tuning} designs a discriminative meta-learning objective to adjust sample weights and demarcate the nuanced differences between different labels.


\subsubsection{Label Enhancement}
{Label enhancement} methods strive to rectify the potentially erroneous annotations in generated samples. Due to confirmation bias, it is unrealistic for LLMs to identify their own mistakes. To address this, recent works either rely on \textit{human intervention} or incorporate a student model for \textit{human-free knowledge distillation}. 

\paragraph{Human Intervention.}  
A straightforward strategy for label refinery is to include human efforts to re-annotate the corrupted samples \cite{chung-etal-2023-increasing,wang-etal-2021-want-reduce,DBLP:journals/corr/abs-2306-00176}. \citet{wang-etal-2021-want-reduce} proposed to actively select samples with the lowest confidence for human re-labeling. \citet{DBLP:journals/corr/abs-2306-00176} and \citet{liu-etal-2022-wanli} further emphasize the importance of human review and suggest comparing annotations from humans and LLMs guided by the same codebook. Despite the simplicity, these methods can lead to considerable labeling costs and can be unrealistic in practical deployment. 

\paragraph{Auxiliary Model.}
To reduce the labeling cost, a more pragmatic human-free paradigm is developed which involves auxiliary student models for knowledge distillation and label refinery \cite{xiao-etal-2023-freeal,DBLP:journals/corr/abs-2311-08640,saad-falcon-etal-2023-udapdr}. These methods rely on the weakly supervised ability of student models and hypothesize that a student distilled from the LLM teacher can produce superior labels. The seminal work FreeAL \cite{xiao-etal-2023-freeal} proposes a collaborative framework, where a student model is leveraged to distill the high-quality task-related knowledge from the weak annotations and in return feedback LLMs for label refinery. MCKD \cite{DBLP:journals/corr/abs-2311-08640} designs a multistage distillation pipeline with data-split training and cross-partition labeling to avoid overfitting on noisy labels. With the expanding abilities and availability of LLMs, the incorporation of auxiliary student models will play a more crucial role as a cost-effective alternative to human intervention.

%% file: sec/5_data_evaluation.tex
\begin{figure}[!ht]
\centering
\includegraphics[width=1\linewidth]{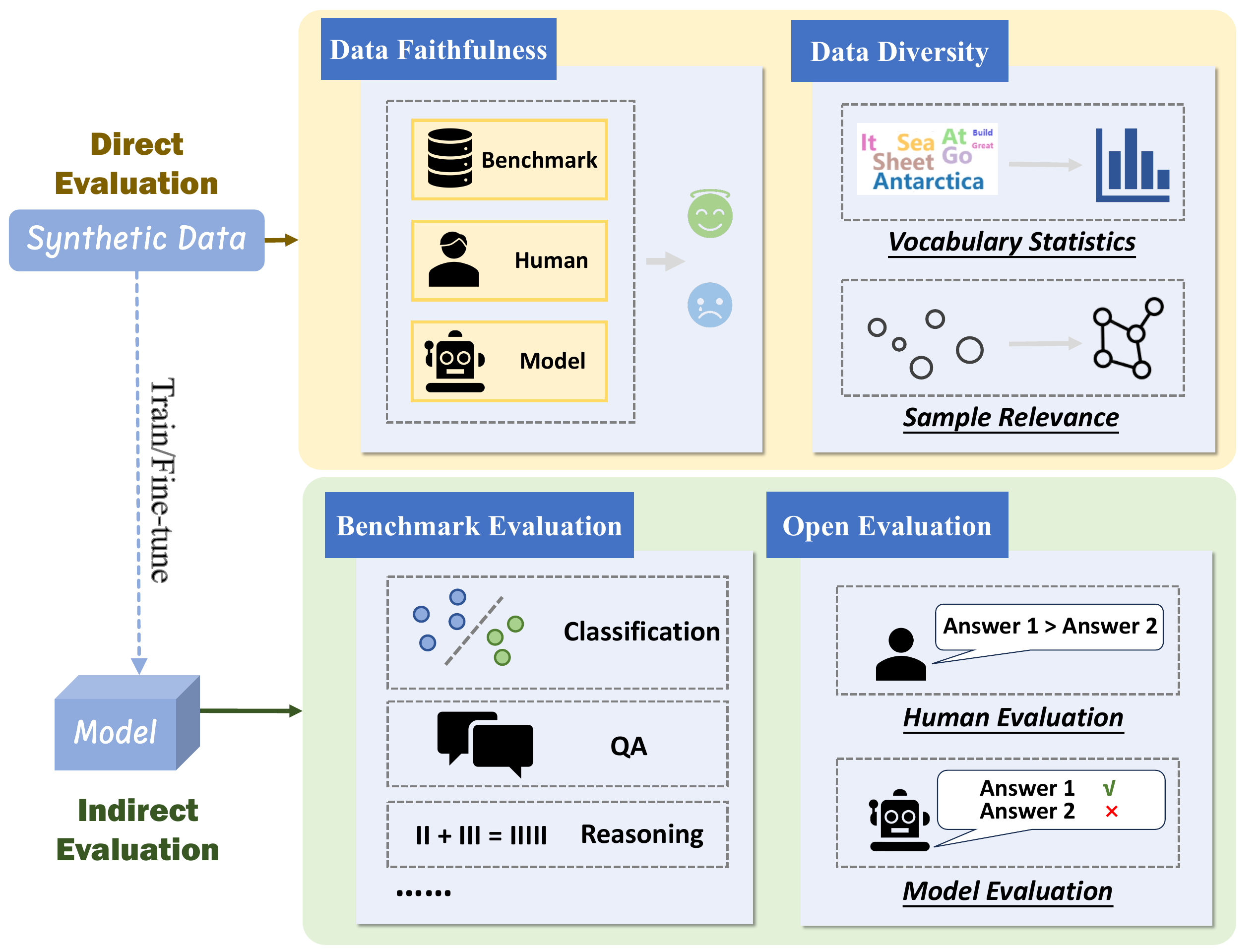}
\caption{Direct and indirect methods of data evaluation.}
\label{fig:evaluation}
\end{figure}

\subsection{Data Evaluation}
Before the employment of generated data, it is important to evaluate the quality and application effectiveness of the data, to ensure its value to downstream tasks. 
The current mainstream evaluation methods can be roughly divided into two categories: \textit{direct} and \textit{indirect}, which evaluate the quality of $\mathcal{D}_\text{gen}$ individually and through its effectiveness on downstream tasks, respectively.

\subsubsection{Direct Evaluation}

\paragraph{Data Faithfulness.}
Ideally, automatic evaluation of the LLMs' generation results can be easily realized with ground truths from existing datasets, if available \cite{abs-2304-10145}. However, for open-ended data, human-based evaluation is necessitated. A straightforward idea is to provide some generated samples to human experts, who will then determine whether they are correct, according to which we can estimate the overall generation quality \cite{WangKMLSKH23}. Theoretically, the larger the sample size, the more accurate the estimation results will be, but the labor it costs will correspondingly get higher. To this end, a reliable auxiliary model can be leveraged for a more comprehensive yet cost-effective evaluation of the generated data in replace of human experts~\cite{DBLP:conf/acl/ChungKA23}. Considering that most models can only process contents of limited length, appropriate information extraction can reduce the burden of the auxiliary model and contribute to a more precise prediction of whether a sample contains factual errors \cite{lee2022factuality}.

\paragraph{Data Diversity.}The quantification of data diversity primarily employs vocabulary statistics and sample relevance calculations. Vocabulary statistics~\cite{abs-2306-15895}, such as vocabulary size and N-gram frequency, provide a straightforward and intuitive approach. However, they struggle to capture the semantic information of a dataset. The calculation of sample relevance compensates for this limitation effectively. The most common measures of sample correlation are based on cosine similarity~\cite{WangZS23} and sample distance~\cite{DBLP:conf/acl/ChungKA23}, which can better capture the contextual and semantic diversity of the dataset. Furthermore, these metrics can also be leveraged to select in-context demonstrations $e_\text{demo}$~\cite{WangKMLSKH23} that are more dissimilar with the previously generated samples, thereby leading to more diversified generation results.

\subsubsection{Indirect Evaluation}
\paragraph{Benchmark Evaluation.}
The performance of downstream models trained on the generated data can also reflect the generation quality to some extent~\cite{abs-2306-15895,DBLP:conf/acl/ChungKA23}. 
Specifically, the impact of synthetic data can be evaluated from multiple dimensions except for the specialized capabilities of the downstream models. For example, TruthfulQA enables the assessment of a model's ability to identify true claims \cite{abs-2305-03047}; NIV2 is employed to evaluate a model's language comprehension and reasoning abilities across multiple tasks \cite{WangKMLSKH23}. 

\paragraph{Open Evaluation.}
For open-ended benchmarks, evaluation by humans or auxiliary models is necessitated due to the absence of standardized answers. To fully leverage the preference outputs of the auxiliary models, multiple evaluation strategies have been designed, such as response ranking~\cite{abs-2304-12244}, four-level rating system~\cite{WangKMLSKH23} and Elo scores~\cite{abs-2212-08073}. To further reduce evaluation costs, \citet{abs-2305-03047,abs-2304-12244} utilize the automatic evaluation framework based on GPT-4 proposed by Vicuna for evaluation. However, general LLMs may lack enough knowledge for domain-specific tasks, which hinders them to provide effective evaluation~\cite{bran2023chemcrow}. Therefore, collecting human assessment data to fine-tune open-source models for evaluation purposes is an important practice in real-world scenarios~\cite{abs-2303-16854}. Other techniques like~\cite{peng2024energy,peng2023came} remain to be further explored.

%% file: sec/7_discussion_conclusion.tex
\section{Future Directions}

\subsection{Complex Task Decomposition}
Current multi-step generation algorithms depend on the model's understanding of task requirements, requiring it to perform complex logical reasoning with limited information. However, in real-world complex scenarios, this limited information may not adequately support effective decision-making.
For instance, the generation of mathematical problem-solution pairs entails multiple reasoning steps and may necessitate the utilization of calculator tools for validation. 
To date, there remains a lack of systematic investigation on how to activate the reasoning and planning capabilities of LLMs for autonomous synthetic data generation. 
Inspired by prevalent LLMs-based agents like HuggingGPT \cite{DBLP:journals/corr/abs-2303-17580} and MetaGPT \cite{DBLP:journals/corr/abs-2308-00352}, we believe it would also be quite valuable to develop a data generation \textit{agent} for industrial applications.

\subsection{Knowledge Enhancement}
Recent research has found that LLMs' knowledge is long-tailed and biased \cite{DBLP:journals/jdiq/NavigliCR23,DBLP:conf/acl/FeiHCB23}. 
Lacking specific domain knowledge, LLMs tend to generate biased, monotonous, and even unfaithful data. 
Though we have introduced how to mildly guide the data generation with task specification and conditional prompting in the previous sections, such methods still hold strong limitations and are not conducive to scalable implementation. 
Instead, we believe that developing automated condition controls directly on mature domain knowledge bases will significantly improve the efficiency of knowledge enhancement. 
For example, we can establish certain links between the LLMs and external knowledge graphs \cite{DBLP:journals/tnn/JiPCMY22} or retrieve augmentation from the website \cite{DBLP:journals/corr/abs-2312-10997}, which is helpful for the definition, decomposition, and reasoning of data features throughout the entire generation process. Additionally, with enhanced domain knowledge, we may also better assess the quality of generated data or even develop automatic evaluation systems. Overall, we believe that knowledge-driven data generation will be a key focus for future studies.

\subsection{Synergy between Large \& Small LMs}
In Section~\ref{sec:data_curation}, we introduced the use of small domain-specific models for data curation. In particular, FreeAL \cite{xiao-etal-2023-freeal} has shown the feasibility of low-cost data curation with integrated collaboration between large and small models. 
The idea of leveraging real-time feedback provided by automated performance evaluation during the data generation process to guide the corresponding adjustments in the following generation hints at an important research direction. 
However, the exploitation of small LMs at the current stage is simply based on prediction confidence. In the future, we are looking forward to seeing more diversified collaboration modes between large and small models to improve the quality of generated data, e.g.,  usage of various output information, new design of collaborative architectures, and so on. 

\subsection{Human-Model Collaboration}
Data, as the source of model intelligence, theoretically cannot be generated completely without human intervention. Otherwise, wild synthetic data that carries noisy, toxic information can easily ``poison'' a model, even resulting in mode collapse. Due to the inherent bias of LLMs, they can hardly be self-aware of the bias in their generated data and finally deviate from our intentions. Thus, designing a human-friendly interactive system to involves a few necessary human knowledge for annotation and verification is vital and irreplaceable. To date, there is still a lack of a generic framework to standardize and systematize the human-machine collaboration involved in the data production process. 

We believe that an appropriate design of such a system must be based on a thorough understanding of the strengths and limitations of human intervention, and should follow the human-centered principle. To achieve sustainable and efficient human involvement, we need comprehensive consideration of various factors such as feasibility, cost, and even labor psychology. For specific examples: (i)-readability and interpretability of the information provided by the LLMs should be ensured to reduce obstacles to human understanding; (ii)-upstream knowledge enrichment or filtering should be carried out to improve the efficiency of human resource utilization and reduce consumption on tasks with low cost-effectiveness; (iii)-incorporating enjoyable interactive features can not only mitigate the negative impact of mechanical data processing tasks on humans but also attract a broader audience.

\section{Conclusion}
In this paper, we present a systematic review of advancements in synthetic data generation propelled by Large Language Models (LLMs). We aim to offer guidance to enterprises and organizations on effectively building their domain-specific datasets using LLMs. In the meantime, we endeavor to provide insights into the challenges and opportunities within this field, while also proposing potential directions for future research.
We hope that our work can promote the rapid production of large amounts of data in various fields and push the limits of data-centric AI. 
We also envision a fantastic future, where an LLMs community, endowed with human-like abilities such as bionics and communication, may be constructed to generate data for its own self-improvement.

%% file: sec/limitations.tex
\section*{Limitations}
In this paper, we survey existing studies on LLMs-driven synthetic data generation, curation, and evaluation, proposing a generic workflow for real-world practice. Synthetic data generation is a broad topic that involves data and models of various modals, including vision and speech. Due to the page limit, we mainly focus on the objective of text data and LLMs-driven approaches, while leaving investigations in other fields for future work. We will also keep paying attention to the latest work and add more related approaches with more detailed analysis.

\section*{Ethics Statement}
We believe that our proposed workflow of LLMs-driven synthetic data generation, curation, and evaluation can benefit both researchers who are interested in data-centric AI and industrial producers who are facing data problems. However, the malicious use of such synthetic data also raises ethical concerns that should arouse our vigilance.

\section*{Acknowledgements}
This work is supported by the Pioneer R\&D Program of Zhejiang (No. 2024C01035), NSFC under Grants (No. 62206247), and the Fundamental Research Funds for the Central Universities  (No. 226-2024-00049).

%% file: sec/appendix.tex
\appendix

\section{Data Annotation} 
\label{app:data_annotation}
In the main text, we introduced a series of techniques for general data synthesis. Though annotation can be considered a special type of synthesis with the input of a particular sample as the synthesis condition, there are also approaches specifically suitable for data annotation. Among them, \textit{selective annotation} is one of the most important practices. Selective annotation represents an optimal tradeoff between expensive and precise human annotation and economic but relatively rough LLMs-based annotation\cite{wang-etal-2021-want-reduce,KoconCKKSBBGJKKKMMOPRWWK23}. 

The key to selective annotation is to define a "cost-effective" sample distribution between humans and LLMs. \cite{zhang-etal-2023-llmaaa,abs-2306-15766} covers some common selection strategies for LLMs-based annotation, including random selection, maximum entropy selection, least confidence selection and $k$means selection for thorough comparisons. 
Results show that uncertainty-based methods, i.e. maximal entropy and least confidence, perform significantly better than the random baseline, with faster convergence and better performance of the downstream model trained on the annotated data. \cite{li-etal-2023-coannotating} also utilizes uncertainty to estimate LLMs’ annotation capability to effectively allocate the annotation work among humans and LLMs.
\cite{SuKWSWX0OZS023} instead proposes a novel unsupervised, graph-based selective annotation method named vote-$k$, to select diverse and representative examples to annotate.

\section{Tuning Techniques}
Another large body of research pertains to the \textit{tuning techniques}, such as model fine-tuning \cite{zhao2023tabula,abs-2305-03047,meng2023tuning,kurakin2023harnessing} and soft prompting \cite{ChenLLRY23}, which have already been heavily studied in other fields and can be detailedly referred in \cite{hu-etal-2023-llm,DBLP:journals/corr/abs-2312-01504,DBLP:conf/bigdataconf/WeiKHZDYMQ23,DBLP:journals/corr/abs-2307-02839}. Despite their effectiveness in improving the generation performance, most of the existing approaches are established on the accessibility of the LLMs, while their application on black-box models remains to be further explored.

\section{Applications}
LLM-driven synthetic data generation has served as a new alternative to traditional human-dependent data collection and demonstrated great potential in various applications, including general tasks, domain-specific tasks, and multimodal tasks. 

\paragraph{Generic Tasks.}
With the exploding capabilities of LLMs, this generation pipeline has been adopted in a wide range of basic NLP studies, including text classification \cite{ye-etal-2022-progen,yu-etal-2023-regen,sahu-etal-2022-data}, named entity recognition \cite{xiao-etal-2023-freeal}, question answering \cite{li-callison-burch-2023-paxqa}, relationship extraction \cite{abs-2303-16854}, and natural language inference \cite{zhang-etal-2023-llmaaa}. These studies further underpin diverse applications, such as sentiment recognition \cite{GaoPLXY0ZLLK23,ye-etal-2022-progen}, online translation \cite{DBLP:journals/corr/abs-2307-16833}, stance detection \cite{li-etal-2023-coannotating} and spam identification \cite{abs-2205-02318}.


\paragraph{Domain-specific Tasks.}
Some domain-specific tasks also impose significant demands on this pipeline, where human annotation can be extremely expensive and impractical, such as medical diagnosis \cite{abs-2303-04360}, drug discovery \cite{xiao-etal-2023-freeal}, clinical trial extraction \cite{abs-2311-00287}, industrial advertisement \cite{DBLP:conf/iclr/ZhangWSW00R22} and tabular data analysis \cite{seedat2023curated}.


\paragraph{Multimodal Tasks.}
Stemming from the simplicity and low cost, this generation paradigm has also exhibited significant promise in multimodal tasks, including text-image retrieval \cite{kritharoula-etal-2023-large}, chat understanding  \cite{DBLP:journals/corr/abs-2311-16483}, visual question answering \cite{han-gardent-2023-multilingual}, and multimodal instruction tuning \cite{DBLP:journals/corr/abs-2310-03744}.



\begin{sidewaystable*}\centering
\resizebox{1\linewidth}{!}{
\begin{tabular}{ccccccc}
\toprule
Type       & Benchmark Dataset           & Subdataset Quantity & Partial Subdataset          & Task & Ability       & Domain/Data Source \\
\midrule
\multirow{10}{*}{Classification} & SMS spam \cite{AlmeidaHY11,LiZL023} & 1  & SMS spam & Text Classification        & Spam Detection   & SMS             \\
        & AG News \cite{ZhangZL15,LiZL023}  & 1  & AG News  & Text Classification        & Topic Classification                & News            \\
           & IMDb \cite{maas-etal-2011-learning,LiZL023,WangZS23}  & 1  & IMDb  & Text Classification        & Binary Sentiment Classification     & Review          \\
           & GoEmotions \cite{demszky-etal-2020-goemotions,LiZL023} & 1  & GoEmotions & Text Classification        & Sentiment Classification            & Reddit Comments \\
           & CLINC150 \cite{LarsonMPCLHKLLT19,sahu-etal-2022-data} & 1  & CLINC150 & Text Classification        & Intent Detection & Human Annotation   \\
           & BANKING77 \cite{casanueva-etal-2020-efficient,sahu-etal-2022-data}  & 1  & BANKING77  & Text Classification        & Intent Detection & Bank            \\
           & FewRel \cite{gao-etal-2019-fewrel,LiZL023}  & 1  & FewRel   & Text Classification        & Relation Classification             & Wikipedia       \\
           & \multirow{2}{*}{GLUE \cite{wang-etal-2018-glue,WangZS23}}      & \multirow{2}{*}{7}  & QNLI  & Natural Language Inference & Recognizing Textual Entailment      & Wikipedia       \\
           &       &  & RTE   & Natural Language Inference & Recognizing Textual Entailment      & News and Wikipedia \\
\midrule
\multirow{2}{*}{QA}             & AdversarialQA \cite{bartolo-etal-2020-beat,WangZS23}  & 1 & AdversarialQA  & Question Answering & Reading Comprehension & Wikipedia \\
           & TruthfulQA \cite{lin-etal-2022-truthfulqa,abs-2305-03047} & 1  &TruthfulQA  & Question Answering         & Honestness    & Hard Data       \\
\midrule
\multirow{2}{*}{Reasoning}  
           & MATH \cite{HendrycksBKABTS21,WanHYQB023}  & 1  & MATH  & mathematical reasoning     & Complex Reasoning  & Math            \\
           & \textbf{ToolBench} \cite{abs-2307-16789}          & \textbf{1}  & \textbf{ToolBench}  & \textbf{Trajectory Planning}        & \textbf{Tool manipulation}  & \textbf{Tool}            \\
\midrule
-          & NIV2 \cite{wang-etal-2022-super,WangKMLSKH23} & 1616                & -     & -    & Language Understanding \& Reasoning & Benchmark Collection/Human Annotation \\
-          & BIG-bench \cite{abs-2206-04615,abs-2305-03047}  & 204                 & -     & -    & Language Understanding \& Reasoning & Human Annotation  \\
\bottomrule
\end{tabular}}
\caption{Representative benchmark dataset for assessing models trained with generated data. The dataset generated based on LLM is highlighted in bold.}\label{tab:benchmark}
\end{sidewaystable*}

\section{Benchmark Datasets}
In Table \ref{tab:benchmark}, we summarize representative benchmark datasets for evaluating models trained through data generation. Among them, ToolBench \cite{abs-2307-16789} is generated by LLMs and is commonly employed to evaluate the performance of LLMs in tool usage proficiency. In most classification task evaluations \cite{LiZL023,WangZS23,sahu-etal-2022-data}, LLMs are infrequently used as test models; instead, small language models trained on generated data are often used, followed by testing on existing benchmarks.